\documentclass[a4paper,twoside]{article}

\usepackage{epsfig}
\usepackage{subfigure}
\usepackage{calc}
\usepackage{amssymb}
\usepackage{amsthm}
\usepackage{multicol}
\usepackage{pslatex}
\usepackage{apalike}
\usepackage[utf8]{inputenc}
\usepackage{times}
\usepackage{latexsym}
\usepackage{graphicx}
\usepackage{amsmath}
\usepackage{float}
\usepackage{a4wide}
\usepackage{todonotes}
\usepackage{cleveref}
\usepackage{glossaries}
\usepackage[]{algorithm2e}
\usepackage{caption}
\usepackage{listings}
\usepackage{hyperref}
\graphicspath{ {images/} }
\usepackage[left=4.00cm, right=4.00cm, top=3.00cm, bottom=3.00cm]{geometry}

\usepackage{SCITEPRESS}     
\begin{document}

\title{Social Emotion Mining Techniques \\for Facebook Posts Reaction Prediction}

\author{Florian Krebs\thanks{Denotes equal contribution}, Bruno Lubascher*, Tobias Moers*, Pieter Schaap*, Gerasimos Spanakis\thanks{Corresponding author}
\affiliation{Department of Data Science and Knowledge Engineering, Maastricht University, Maastricht, Netherlands}
Email: \email{\{florian.krebs, bruno.lubascher, tobias.moers, pieter.schaap\}@student.maastrichtuniversity.nl,\\ jerry.spanakis@maastrichtuniversity.nl}}



\abstract{
As of February 2016 Facebook   allows users to express their experienced emotions about a post by using five so-called `reactions'. This research paper proposes and evaluates alternative methods for predicting these reactions to user posts on public pages of firms/companies (like supermarket chains). For this purpose, we collected posts (and their reactions) from Facebook pages of large supermarket chains and constructed a dataset which is available for other researches. In order to predict the distribution of reactions of a new post, neural network architectures (convolutional and recurrent neural networks) were tested using pretrained word embeddings. Results of the neural networks were improved by introducing a bootstrapping approach for sentiment and emotion mining on the comments for each post. The final model (a combination of neural network and a baseline emotion miner) is able to predict the reaction distribution on Facebook posts with a mean squared error (or misclassification rate) of 0.135.
}

\onecolumn \maketitle \normalsize \vfill

\setcounter{footnote}{0}

\section{\uppercase{Introduction}}
The ability to accurately classify the sentiment of short sentences such as Facebook posts or tweets is essential to natural language understanding. In recent years, more and more users share information about their customer experience on social media pages related to (and managed by) the equivalent firms/companies. Generated data attracts a lot of research towards sentiment analysis with many applications in political science, social sciences, business, education, etc. \cite{ortigosa2014sentiment}, \cite{feldman2013techniques}, \cite{troussas2013sentiment}.

Customer experience (CX) represents a holistic perspective on customer encounters with a firm’s products or services. Thus, the more managers can understand about the experiences customers have with their product and service offerings, the more they can measure them again in the future to influence purchase decisions. The rise of social media analytics \cite{fan2014power} offers managers a tool to manage this process with customer opinion data being widely available on social media. Analysing Facebook posts can help firm managers to better manage posts by allowing customer care teams to reply faster to unsatisfied customers or maybe even delegate posts to employees based on their expertise. Also, it would be possible to estimate how the reply on a post affects the reaction from other customers. To our knowledge, no previous research work on predicting Facebook reaction posts exists.

The main goals and contributions of this paper are the following: (a) contribute a dataset which can be used for predicting reactions on Facebook posts, useful for both machine learners and marketing experts and (b) perform sentiment analysis and emotion mining to Facebook posts and comments of several supermarket chains by predicting the distribution of the user reactions. Firstly, sentiment analysis and emotion mining baseline techniques are utilized in order to analyse the sentiment/emotion of a post and its comments. Afterwards, neural networks with pretrained word embeddings are used in order to accurately predict the distribution of reactions to a post. Combination of the two approaches gives a working final ensemble which leaves promising directions for future research.

The remainder of the paper is organized as follows. Section \ref{sec:rw} presents related work about sentiment and emotion analysis on short informal text like from Facebook and Twitter. The used dataset is described in Section \ref{sec:dataset}, followed by the model (pipeline) description in Section \ref{sec:predictionsystem}. Section \ref{sec:experiments} presents the experimental results and finally, Section \ref{sec:conclusion} concludes the paper and presents future research directions. 

\section{\uppercase{Related Work}}
\label{sec:rw}
Deep learning based approaches have recently become more popular for sentiment classification since they automatically extract features based on word embeddings. Convolutional Neural Networks (CNN), originally proposed in \cite{lecun1998gradient} for document recognition, have been extensively used for short sentence sentiment classification. \cite{Kim14f} uses a CNN and achieves state-of-the art results in sentiment classification. They also highlight that one CNN layer in the model's architecture is sufficient to perform well on sentiment classification tasks. Recurrent Neural Networks (RNN) and more specifically their variants Long Short Term Memory (LSTM) networks \cite{hochreiter1997long} and Gated Recurrent Units (GRU) networks \cite{chung2014empirical} have also been extensively used for sentiment classification since they are able to capture long term relationships between words in a sentence while avoiding vanishing and exploding gradient problems of normal recurrent network architectures \cite{hochreiter1998vanishing}. \cite{wang2014sentiment} proves that combining different architectures, such as CNN and GRU, in an ensemble learner improves the performance of individual base learners for sentiment classification, which makes it relevant for this research work as well. 

Most of the work on short text sentiment classification concentrates around Twitter and different machine learning techniques \cite{wang2011topic}, \cite{kouloumpis2011twitter}, \cite{saif2012semantic}, \cite{sarlan2014twitter}. These are some examples of the extensive research already done on Twitter sentiment analysis. Not many approaches for Facebook posts exist, partly because it is difficult to get a labeled dataset for such a purpose. 

Emotion lexicons like EmoLex \cite{emolex} can be used in order to annotate a corpus, however, results are not satisfactory and this is the reason that bootstrapping techniques have been attempted in the past. For example, \cite{bootstrap} propose such a technique which enhances EmoLex with synonyms and then combines word vectors \cite{mikolov2013efficient} in order to annotate more examples based on sentence similarity measures. 

Recently, \cite{tian2017facebook} presented some first results which associate Facebook reactions with emojis but their analysis stopped there. \cite{pool2016distant} utilized the actual reactions on posts in a distant supervised fashion to train a support vector machine classifier for emotion detection but they are not attempting at actually predicting the distribution of reactions.

Moreover, analysis of customer feedback is an area which gains interest for many companies over the years. Given the amount of text feedback available, there are many approaches around this topic, however none of them are handling the increasing amounts of information available through Facebook posts. For the sake of completeness, we highlight here some these approaches. Sentiment classification (\cite{pang2002thumbs}, \cite{glorot2011domain}, \cite{socher2013recursive}) deals only with the sentiment analysis (usually mapping sentiments to positive, negative and neutral (or other 5-scale classification) and similarly emotion classification (\cite{yang2007emotion}, \cite{wen2014emotion} only considers emotions. Some work exists on Twitter data \cite{pak2010twitter} but does not take into account the reactions of Facebook. Moreover, work has been conducted towards customer review analysis (\cite{yang2004online}, \cite{hu2004mining}, \cite{cambria2013new}) but none of them are dealing with the specific nature of Facebook (or social media in general). 

In this work, we combine sentiment analysis and emotion mining techniques with neural network architectures in order to predict the distribution of reactions on Facebook posts and actually demonstrate that such an approach is feasible.

\section{\uppercase{Dataset Construction}}
\label{sec:dataset}
\begin{figure}[h!]
	\centering
	\scalebox{0.9}{
	\includegraphics[width=\columnwidth]{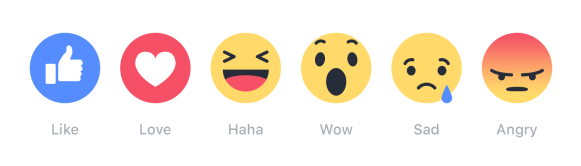}}
	\caption{The Facebook reaction icons that users are able to select for an original post.}
	\label{fig:fbreactions}
\end{figure}

Our dataset consists out of Facebook posts on the customer service page of 12 US/UK big supermarket/retail chains, namely Tesco, Sainsbury, Walmart, AldiUK, The Home Depot, Target, Walgreens, Amazon, Best Buy, Safeway, Macys and publix. The vast majority of these posts are initiated by customers of these supermarkets. In addition to the written text of the posts, we also fetch the Facebook's reaction matrix \footnote{\url{http://newsroom.fb.com/news/2016/02/reactions-now-available-globally/}} as well as the comments attached to this post made by other users. Such reactions only belong to the initial post, and not to replies to the post since the feature to post a reaction on a reply has only been introduced very recently (May 2017) and would result in either a very small dataset or an incomplete dataset. These reactions include \textit{like}, \textit{love}, \textit{wow}, \textit{haha}, \textit{sad}, \textit{angry} as shown in Figure \ref{fig:fbreactions}. This form of communication was introduced by Facebook on February 24th, 2016 and allows users to express an `emotion' towards the posted content.

\begin{figure}[!h]
	    \centering
	    \scalebox{1}{
	    \includegraphics[width = \columnwidth]{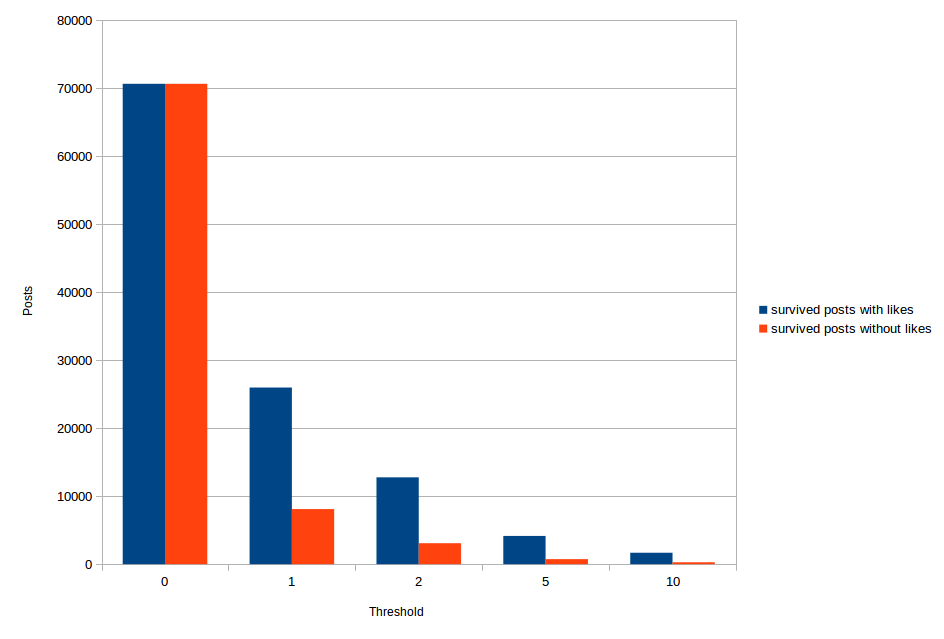}}
	    \caption{Amount of survived posts for different thresholds including/excluding likes}
	    \label{fig:total_post_thresholds}
    \end{figure}

In total, there were more than 70,000 posts without any reaction, thus they were excluded from the dataset. Apart from this problem, people are using the `like' reaction not only to show that they like what they see/read but also to simply tell others that they have seen this post or to show sympathy. This results in a way too often used `like'-reaction which is why likes could be ignored in the constructed dataset. So, instead of using all crawled data, the developed models will be trained on posts that have at least one other reaction than likes. After applying this threshold the size of the training set reduced from 70,649 to 25,969. The threshold of 1 is still not optimal since it leaves much space for noise in the data (e.g. miss-clicked reactions) but using a higher threshold will lead to extreme loss of data. Statistics on the dataset and on how many posts `survive' by using different thresholds can be seen in Figure \ref{fig:total_post_thresholds}.

\begin{figure}[h!]
	\centering
	\scalebox{1}{
	\includegraphics[width=\columnwidth]{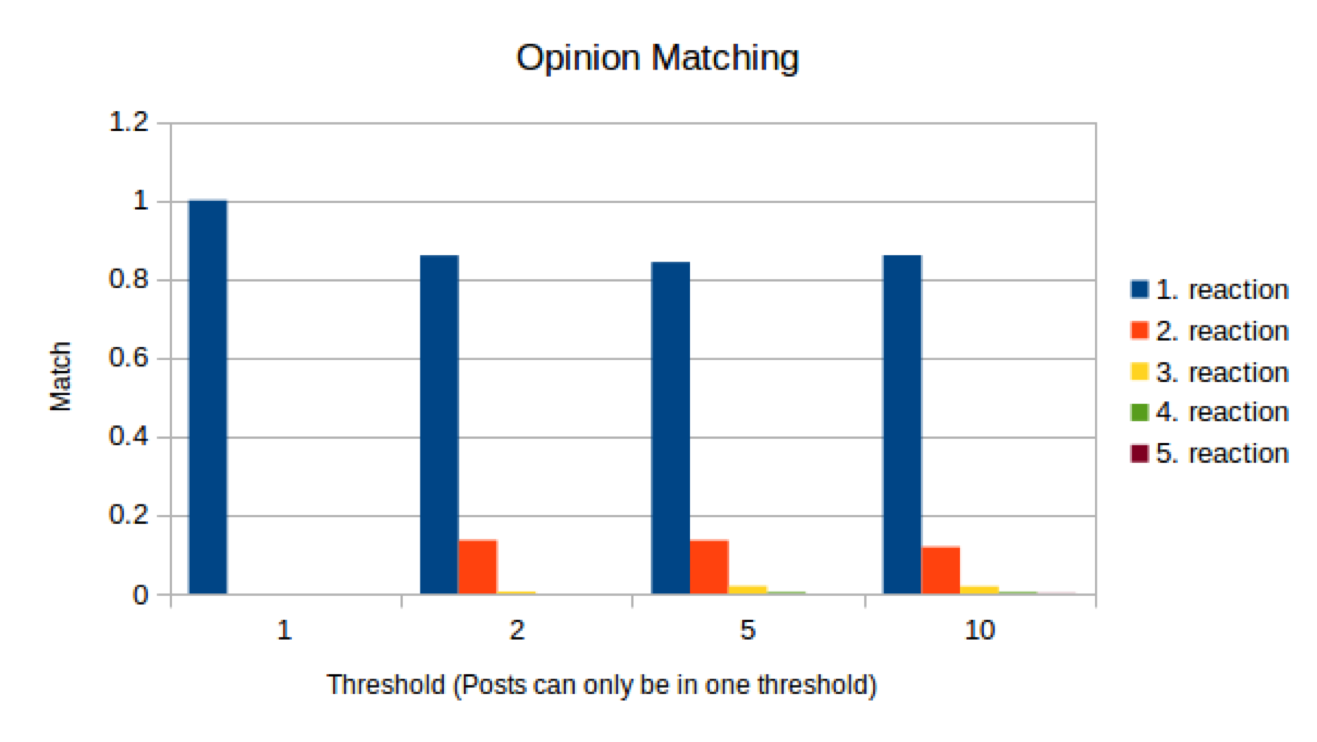}}
	\caption{Reaction match when there is more than one type}
	\label{fig:react}
\end{figure}

\begin{figure}[h!]
	\centering
	\scalebox{1}{
	\includegraphics[width=\columnwidth]{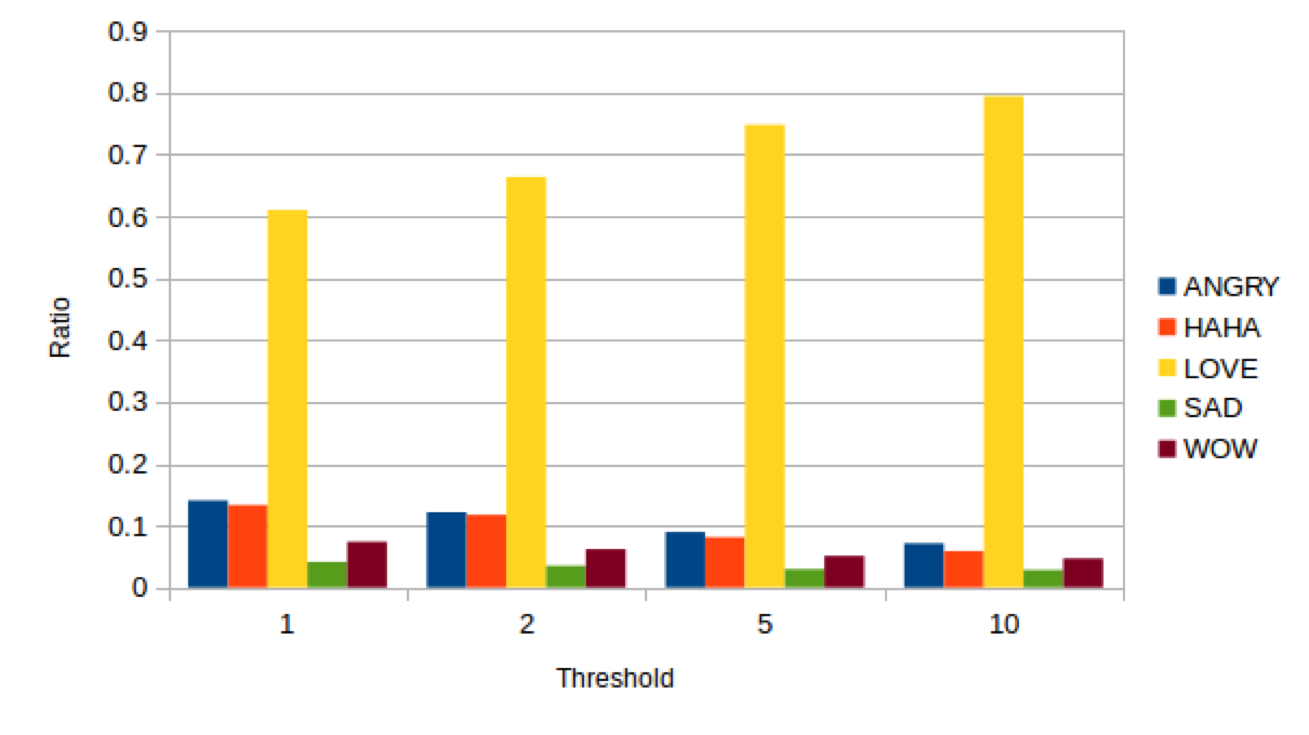}}
	\caption{Distribution of reactions with different minimum thresholds}
	\label{fig:equal}
\end{figure}

Exploratory analysis on the dataset shows that people tend to agree in the reactions they have to Facebook posts (which is consistent for building a prediction system), i.e. whenever there are more than one types of reactions they seem to be the same in a great degree (over 80 \%) as can be seen in Figure \ref{fig:react}. In addition, Figure \ref{fig:equal} shows that even by excluding the \textit{like} reaction, which seems to dominate all posts, the distribution of the reactions remains the same, even if the threshold of minimum reactions increases. Using all previous insights and the fact that there are 25,969 posts with at least one reaction and since the \textit{like} reaction dominates the posts, we chose to include posts with at least one reaction which is not a \textit{like}, leading to finally 8,103 posts. Full dataset is available \footnote{\url{https://github.com/jerryspan/FacebookR}} and will be updated as it is curated and validated at the moment of the paper submission.

\subsection{Pre-processing}
Pre-processing on the dataset is carried out using the Stanford CoreNLP parser \cite{corenlp} and includes the following steps:
\begin{itemize}
\item Convert everything to lower case
\item Replace URLs with ``\_\_URL\_\_" as a generic token
\item Replace user/profile links with ``\_\_AT\_USER\_\_" as a generic token
\item Remove the hash from a hashtag reference (e.g. \#hashtag becomes ``hashtag")
\item Replace three or more occurrences of one character in a row with the character
itself (e.g. ``looooove" becomes "love")
\item Remove sequences containing numbers (e.g. ``gr34t")
\end{itemize}

Afterwards, each post is split using a tokenizer based on spaces and after some stop-word filtering the final list of different tokens is derived. Since pre-processing on short text has attracted much attention recently \cite{singh2016role}, we also demonstrate the effect of it on the developed models in the Experiments section.

\section{\uppercase{Reaction distribution prediction system pipeline}}
\label{sec:predictionsystem}
In this Section, the complete prediction system is described. There are three core components: emotion mining applied to Facebook comments, artificial neural networks that predict the distribution of the reactions for a Facebook post and a combination of the two in the final prediction of the distribution of reactions.

\subsection{Emotion mining}
\label{sec:emotionmining}
\begin{figure*}[h!]
	\centering
	\scalebox{0.9}{
	\includegraphics[width=\textwidth]{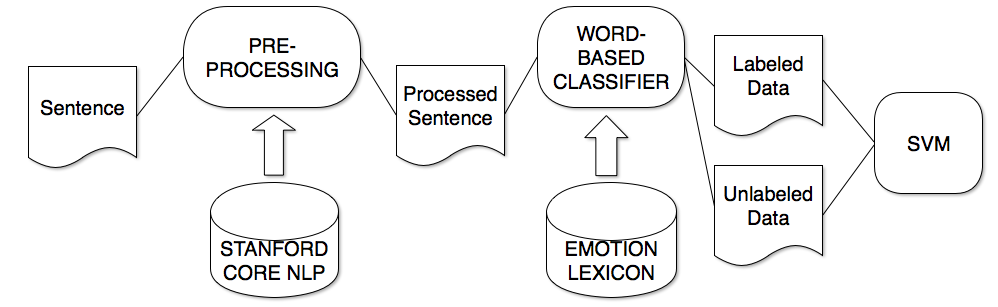}}
	\caption{Emotion miner pipeline}
	\label{fig:miner}
\end{figure*}

The overall pipeline of the emotion miner can be found in Figure \ref{fig:miner}.

The emotion lexicon that we utilize is created by \cite{emolex} and is called NRC Emotion Lexicon (EmoLex). This lexicon consists of 14,181 words with eight basic emotions (anger, fear, anticipation, trust, surprise, sadness, joy, and disgust) associated with each word in the lexicon. It is possible that a single word is associated with more than one emotion. An example can be seen in Table \ref{tab:emolex}. Annotations were manually performed by crowd-sourcing.

\begin{table}[H]
	\caption{Examples from EmoLex showing the emotion association to the words abuse and shopping.}
    \resizebox{\columnwidth}{!}{%
	\begin{tabular}{|l|l|l|l|l|l|l|l|l|}
		\hline
		 & \textbf{Anger} & \textbf{Anticipation} & \textbf{Disgust} & \textbf{Fear} & \textbf{Joy}& \textbf{Sadness}& \textbf{Surprise}& \textbf{Trust} \\
		\hline
		\textbf
		abuse & \textbf{1} & 0 & \textbf{1}& \textbf{1} & 0 & \textbf{1}& 0 & 0 \\
		\hline
		shopping & 0 & \textbf{1} & 0 & 0 & \textbf{1} & 0 & \textbf{1} & \textbf{1} \\
		\hline
	\end{tabular}}
	\label{tab:emolex}
\end{table}

Inspired by the approach of \cite{bootstrap}, EmoLex is extended by using WordNet \cite{wordnet}: for every synonym found, new entries are introduced in EmoLex having the same emotion vector as the original words. By applying this technique the original database has increased in size from 14,181 to 31,485 words that are related to an emotion vector. The lexicon can then be used to determine the emotion of the comments to a Facebook post. For each sentence in a comment, the emotion is determined by looking up all words in the emotion database and the found emotion vectors are added to the sentence emotion vector. By merging and normalizing all emotion vectors, the final emotion distribution for a particular Facebook post, based on the equivalent comments, can be computed. However, this naive approach yielded poor results, thus several enhancements were considered, implemented and described in subsections \ref{sec:negation}-\ref{sec:svm}.

\subsubsection{Negation Handling}
\label{sec:negation}
The first technique that was used to improve the quality of the mined emotions is negation handling. By detecting negations in a sentence, the ability to `turn' this sentiment or emotion is provided. In this paper only basic negation handling is applied since the majority of the dataset contains only small sentences and this was proved to be sufficient for our goal. The following list of negations and pre- and suffixes are used for detection (based on work of \cite{neg}):

\begin{table}[h]
  \centering
  \caption{Negation patterns}
  \resizebox{\columnwidth}{!}{%
    \begin{tabular}{|c|p{2.0in}|}
    \hline
    Negations & no, not, rather, wont, never, none, nobody, nothing, neither, nor, nowhere, cannot, without, n't \\
        \hline
    Prefixes & a, de, dis, il, im, in, ir, mis, non, un \\
    \hline
    Suffixes & less \\
    \hline
    \end{tabular}}
  \label{tab:negations}%
\end{table}%

The following two rules are applied:
\begin{enumerate}
	\item The first rule is used when a negation word is instantly followed by an emotion-word (which is present in our emotion database).
	\item The second rule tries to handle adverbs and past particle verbs (Part-of-Speech (POS) tags: RB, VBN).
	If a negation word is followed by one or more of these POS-tags and a following emotion-word, the emotion-word's value will be negated.
	For example this rule would apply to `not very happy'.
\end{enumerate}

\noindent There are two ways to obtain the emotions of a negated word:
\begin{enumerate}
	\item Look up all combinations of negation pre- and suffixes together with the word in our emotion lexicon.
	\item If there is no match in the lexicon a manually created mapping is used between the emotions and their negations.
	This mapping is shown in Table \ref{tab:mappingemotions}.
\end{enumerate}

\begin{table}[H]
	\caption{Mapping between emotion and negated emotions.}
    \resizebox{\columnwidth}{!}{%
	\begin{tabular}{|l|l|l|l|l|l|l|l|l|}
		\hline
		  & \textbf{Anger} & \textbf{Anticipation} & \textbf{Disgust} & \textbf{Fear} & \textbf{Joy}& \textbf{Sadness}& \textbf{Surprise}& \textbf{Trust} \\
		\hline
		\textbf{Anger} & 0 & 0 & 0 & 0 & \textbf{1}& 0 & 0 & 0 \\
		\hline
		\textbf{Anticipation}& 0 & 0 & 0 & 0 & \textbf{1}& 0 & \textbf{1} & 0 \\
		\hline 
		\textbf{Disgust} & 0 & 0 & 0 & 0 & \textbf{1}& 0 & 0 & \textbf{1} \\
		\hline 
		\textbf{Fear} & 0 & 0 & 0 & 0 & \textbf{1}& 0 & 0 & \textbf{1} \\
		\hline
		\textbf{Joy} & \textbf{1} & 0 & \textbf{1} & \textbf{1 }& 0& \textbf{1} & 0 & 0 \\
		\hline 
		\textbf{Sadness} & 0 & 0 & 0 & \textbf{1} & 0& 0 & 0 & 0 \\
		\hline
		\textbf{Surprise} & 0 & \textbf{1} & 0 & 0 & 0& 0 & 0 &\textbf{1} \\
		\hline
		\textbf{Trust} & 0 & 0 &\textbf{1} & 0 & 0& 0 & \textbf{1} & 0 \\
		\hline
	\end{tabular}}

	\label{tab:mappingemotions}
\end{table}

\subsubsection{Sentence similarity measures}
\cite{bootstrap}'s approach is using word vectors \cite{mikolov2013efficient} in order to calculate similarities between sentences and further annotate sentences. In the context of this paper, a more recent approach was attempted \cite{sentence2vec}, together with an averaging word vector approach for comparison. \cite{sentence2vec} creates a representation for a whole sentence instead of only for one word as word2vec. The average word vector approach is summing up the word vector of each word and then taking the mean of this sum. To find a similarity between two sentences, one then uses the cosine similarity. Surprisingly, both approaches return comparable similarity scores. One main problem which occurred here is that two sentences with different emotions but with the same structure are measured as `similar'. This problem is exemplified with an example:

\begin{verbatim}
Sentence 1: "I really love your car."
Sentence 2: "I really hate your car."
Sentence2Vec similarity: 0.9278
Avg vector similarity: 0.9269
\end{verbatim}
\noindent

This high similarity is problematic since the emotions of the two sentences are completely different. Also, one can see that the two models output almost the same result and that there is no advantage by using the approach of \cite{sentence2vec} over the simple average word vector approach. Hence, the sentence similarity measure method to annotate more sentences is not suited for this emotion mining task because one would annotate positive emotions to a negative sentence and was not adapted for further use.

\subsubsection{Classification of not annotated sentences}
\label{sec:svm}
\begin{figure}[h!]
	\centering
	\scalebox{1}{
	\includegraphics[width=\columnwidth]{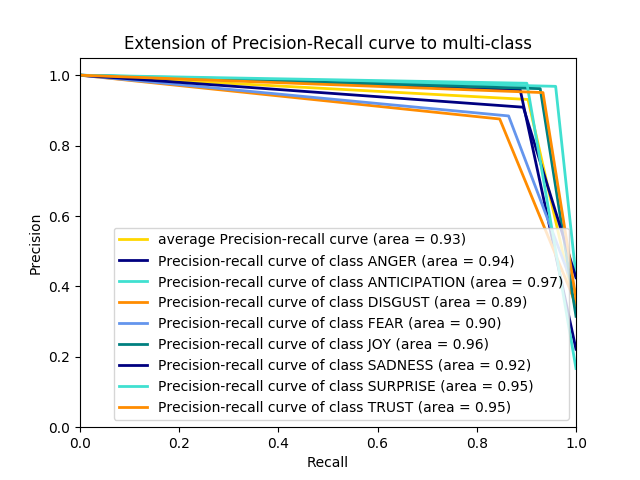}}
	\caption{Precision-Recall (ROC) curve using a linear SVM in an one-versus-all classifier}
	\label{fig:precisionrecall}
\end{figure}

If after performing these enhancement steps there remain any non-emotion-annotated sentences, then a Support Vector Machine (SVM) is used to estimate the emotions of these sentences based on the existing annotations. The SVM is trained as a one-versus-all classifier with a linear kernel (8 models are trained, one for each emotion of EmoLex) and the TF-IDF model \cite{salton1988term} is used for providing the input features. Input consists of a single sentence as data (transformed using the TF-IDF model) and an array of 8 values representing the emotions as a label. With a training/test-split of 80\%/20\%, the average precision-recall is about 0.93. Full results of the SVM training can be seen in Figure \ref{fig:precisionrecall} together with the precision-recall curve for all emotions. The result in this case was judged to be satisfactory in order to utilize it for the next step, which is the reaction prediction and is used as presented here.

\subsection{Reaction distribution predictor}
In order to predict the distribution of the post reactions, neural networks are built and trained using Tensorflow \cite{abadi2016tensorflow}. Two networks were tested, based on literature research: a Convolutional Neural Network (CNN) and a Recurrent Neural Network (RNN) that uses LSTMs.
    
Both networks start with a word embedding layer. Since the analysed posts were written in English, the GloVe \cite{pennington2014glove} pretrained embeddings (with 50 as a vector dimension) were used. Moreover, posts are short texts and informal language is expected, thus we opted for using embeddings previously trained on Twitter data instead of the Wikipedia versions.
    
\subsubsection{CNN}
\begin{figure}[h!]
	\centering
	\scalebox{1}{
	\includegraphics[width=\columnwidth]{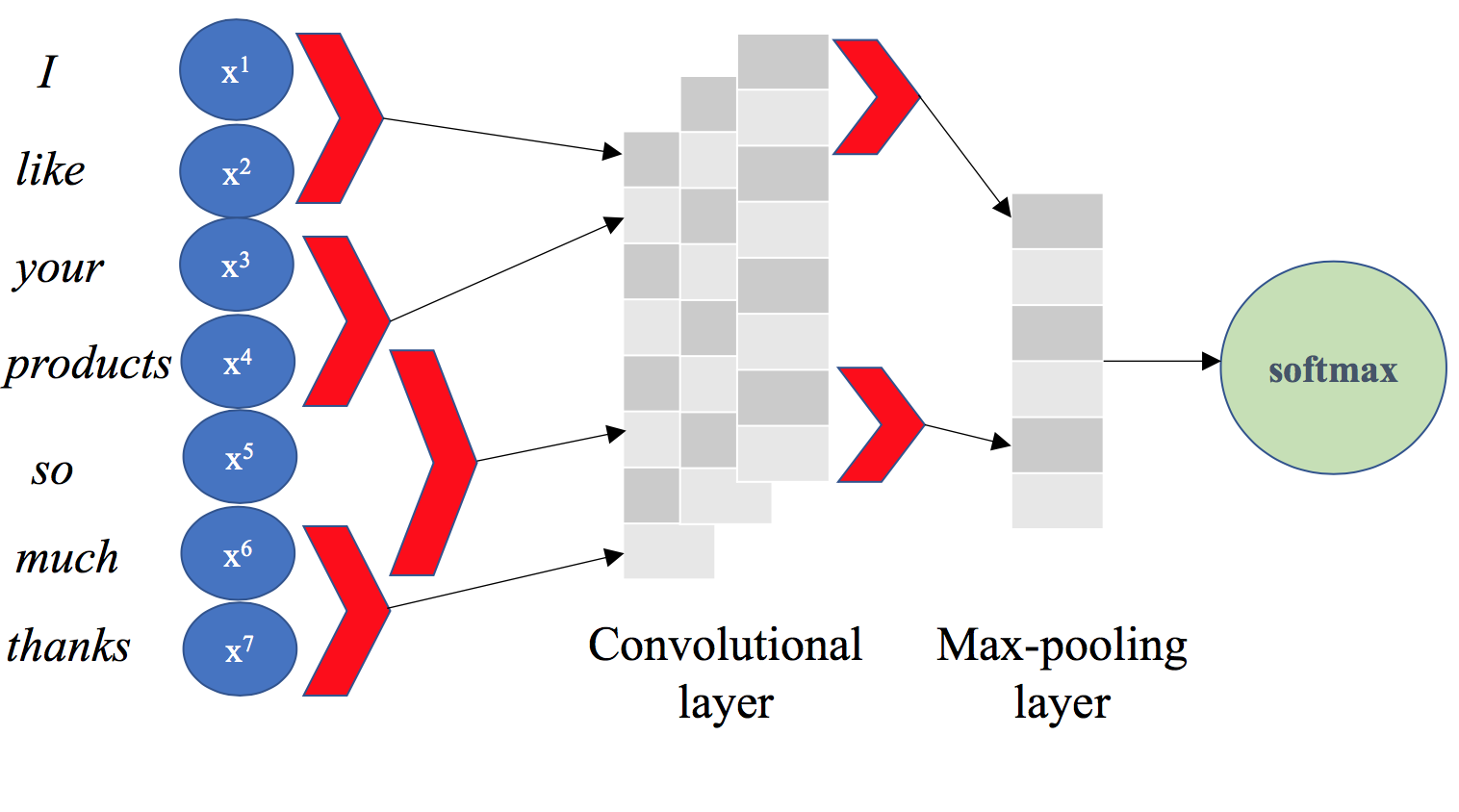}}
	\caption{Convolutional network architecture example}
	\label{fig:cnn1}
\end{figure}

The CNN model is based on existing successful architectures (see \cite{Kim14f}) but is adapted to give a distribution of reactions as an output. An overview of the used architecture is provided in Figure \ref{fig:cnn1}. 

First issue to be handled with CNNs is that since they deal with variable length input sentences, padding is needed so as to ensure that all posts have the same length. In our case, we padded all posts to the maximum post length which also allows efficient batching of the data. In the example of Figure \ref{fig:cnn1} the length of the sentence is 7 and each word $x_i$ is represented by the equivalent word vector (of dimension 50).

The convolutional layer is the core building block of a CNN. Common patterns in the training data are extracted by applying the convolution operation which in our case is limited into 1 dimension: we adjust the height of the filter, i.e. the number of adjacent rows (words) that are considered together (see also red arrows in Figure \ref{fig:cnn1}). These patterns are then fed to a pooling layer. The primary role of the pooling layer is to reduce the spatial dimensions of the learned representations (that's why this layer is also known to perform downsampling). This is beneficial, since it controls for over-fitting but also allows for faster computations. Finally, the output of the pooling layer is fed to a fully-connected layer (with dropout) which has a softmax as output and each node corresponds to each predicted reaction (thus we have six nodes initially). However, due to discarding \textit{like} reaction later on in the research stage, the effective number of output nodes was decreased to 5 (see Experiments). The softmax classifier computes a probability distribution over all possible reactions, thus provides a probabilistic and intuitive interpretation.

\subsubsection{RNN}
\begin{figure}[h!]
	\centering
	\scalebox{1}{
	\includegraphics[width=\columnwidth]{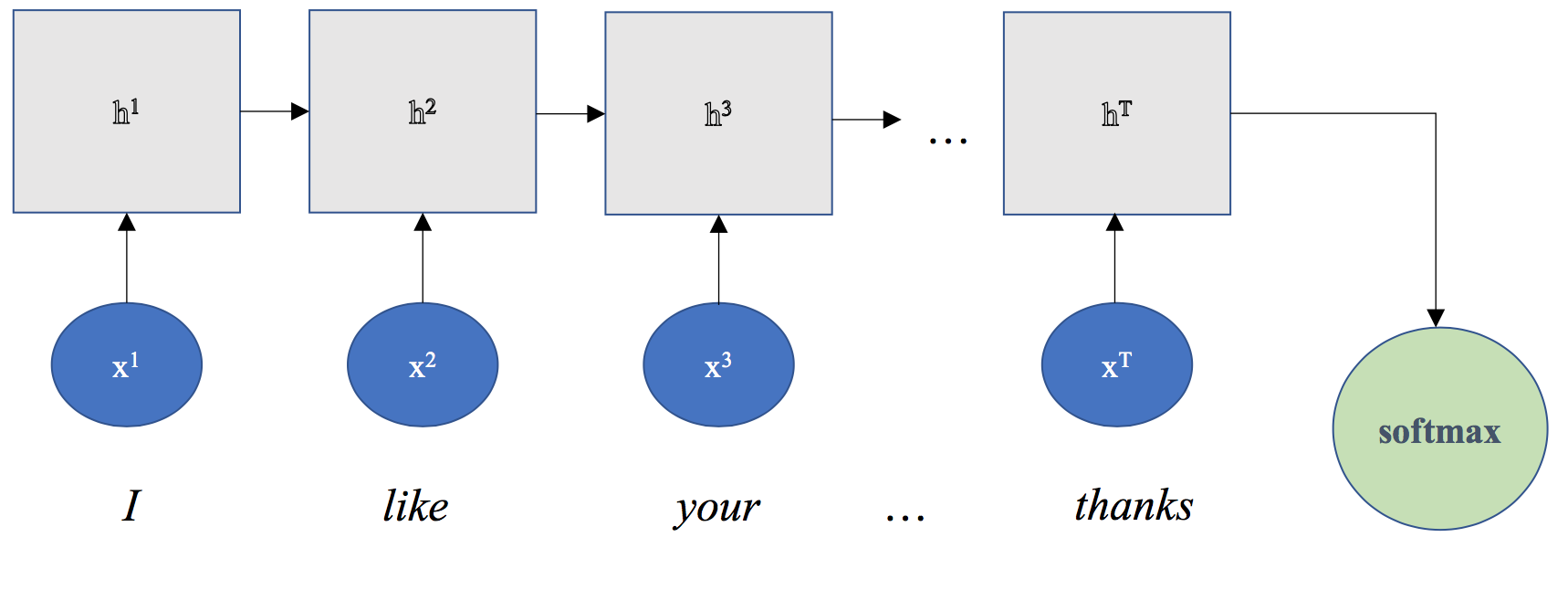}}
	\caption{Recurrent network architecture example}
	\label{fig:rnn1}
\end{figure}

Long short-term memory networks (LSTM) were proposed by \cite{hochreiter1997long} in order to adress the issue of learning long-term dependencies. The LSTM maintains a separate memory cell inside it that updates and exposes its content only when deemed necessary, thus making it possible to capture content as needed. The implementation used here is inspired by \cite{graves2013generating} and an overivew is provided in Figure \ref{fig:rnn1}.

An LSTM unit (at each time step $t$) is defined as a collection of vectors: the input gate ($i_t$), the forget gate ($f_t$), the output gate ($o_t$), a memory cell ($c_t$) and a hidden state ($h_t$). Input is provided sequentially in terms of word vectors ($x_t$) and for each time step $t$ the previous time step information is used as input. Intuitively, the forget gate controls the amount of which each unit of the memory cell is replaced by new info, the input gate controls how much each unit is updated, and the output gate controls the exposure of the internal memory state.

In our case, the RNN model utilizes one recurrent layer (which has 50 LSTM cells) and the rest of the parameters are chosen based on current default working architectures. The output then comes from a weighted fully connected 6-(or 5, depending on the number of reactions)-class softmax layer. Figure \ref{fig:rnn1} explains the idea of recurrent architecture based on an input sequence of words.

\subsection{Prediction ensemble}
\begin{figure*}[h!]
	\centering
	\scalebox{0.8}{
	\includegraphics[width=\textwidth]{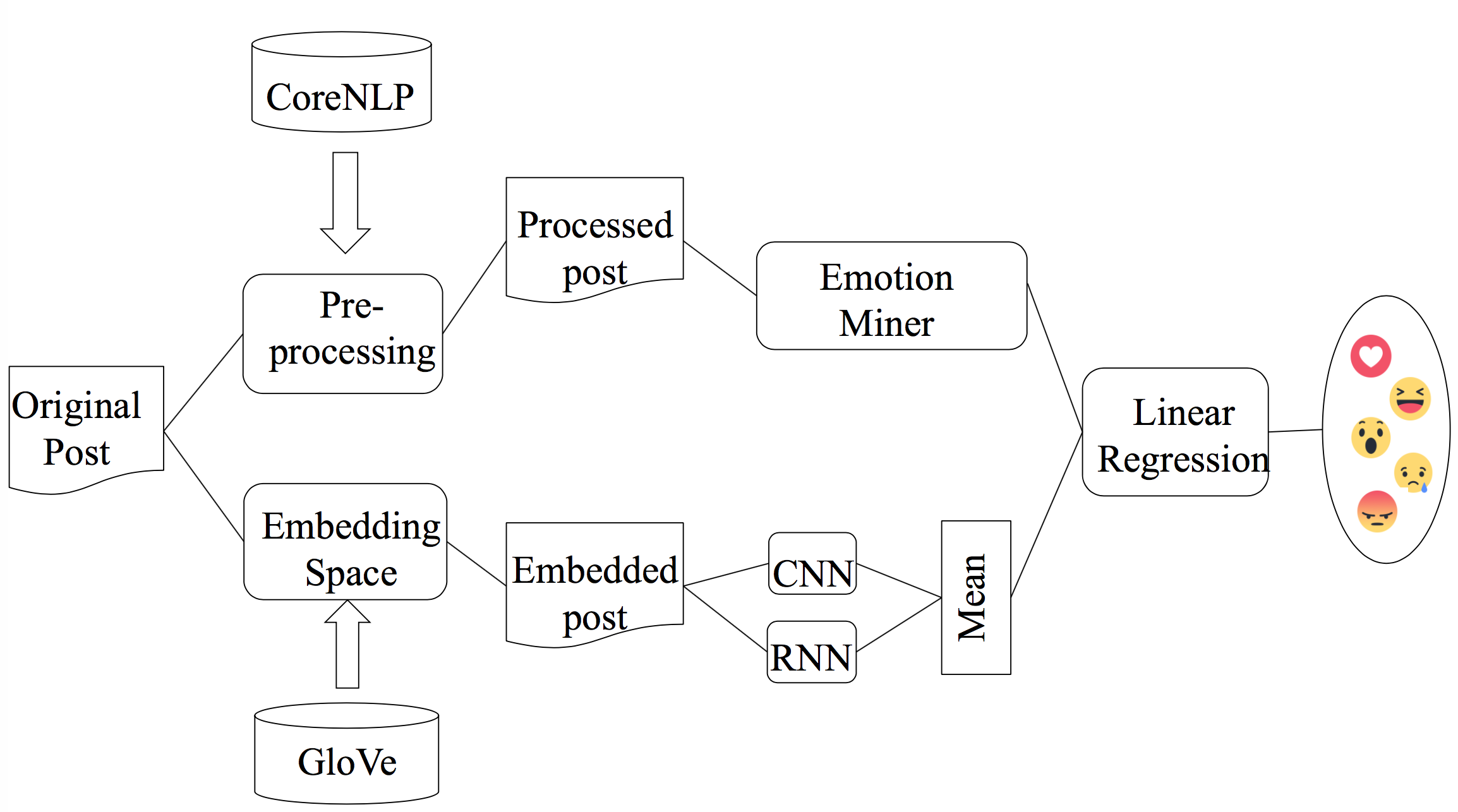}}
	\caption{Pipeline for final prediction of reaction distributions}
	\label{fig:pipeline}
\end{figure*}

The final reaction ratio prediction is carried out by a combination of the neural networks and the mined emotions on the post/comments. For a given post, both networks provide an estimation of the distributions, which are then averaged and normalised. Next, emotions from the post and the comments are extracted following the process described in Section \ref{sec:emotionmining}. The ratio of estimations and emotions are combined into a single vector which is then computed through a simple linear regression model, which re-estimates the predicted reaction ratios. The whole pipeline combining the emotion miner and the neural networks can be seen in Figure \ref{fig:pipeline} and experimental results are presented in the next Section.

\section{\uppercase{Experiments}}
\label{sec:experiments}
Several experiments were conducted in order to assess different effects on the reaction distribution prediction. Firstly, the effect of pre-processing on posts is examined in subsection \ref{sec:preprocessing}. Since Facebook reactions were not introduced too long ago, a lot of posts in the dataset still contain primarily \textit{like} reactions. This might lead to uninteresting results as described in the Dataset Section and in Subsection \ref{sec:exLikes}. Finally, Subsection \ref{sec:mse} discusses the training with respect to the mean squared error (MSE) for CNN and RNN models, as well as the effect of the ensembled approach. 

As mentioned before, both networks utilized the GloVe pre-trained embeddings (with size 50). Batch size was set to 16 for the CNN and 100 for the RNN/LSTM. 

CNN used 40 filters for the convolution (with varying height sizes from 3 to 5), stride was set to 1 and padding to the maximum post length was used. Rectified Linear Unit (ReLU) \cite{glorot2011deep} activation function was used.

Learning rate was set to 0.001 and dropout was applied to both networks and performance was measured by the cross entropy loss with scores and labels with L2-regularization \cite{masnadi2009design}. Mean Squared Error (MSE) is used in order to assess successful classifications (which effectively means that every squared error will be a 1) and in the end MSE is just the misclassification rate of predictions.

\subsection{Raw vs Pre-processed Input}
\label{sec:preprocessing}
\begin{figure}[h]
    \centering
     \includegraphics[width=\columnwidth]{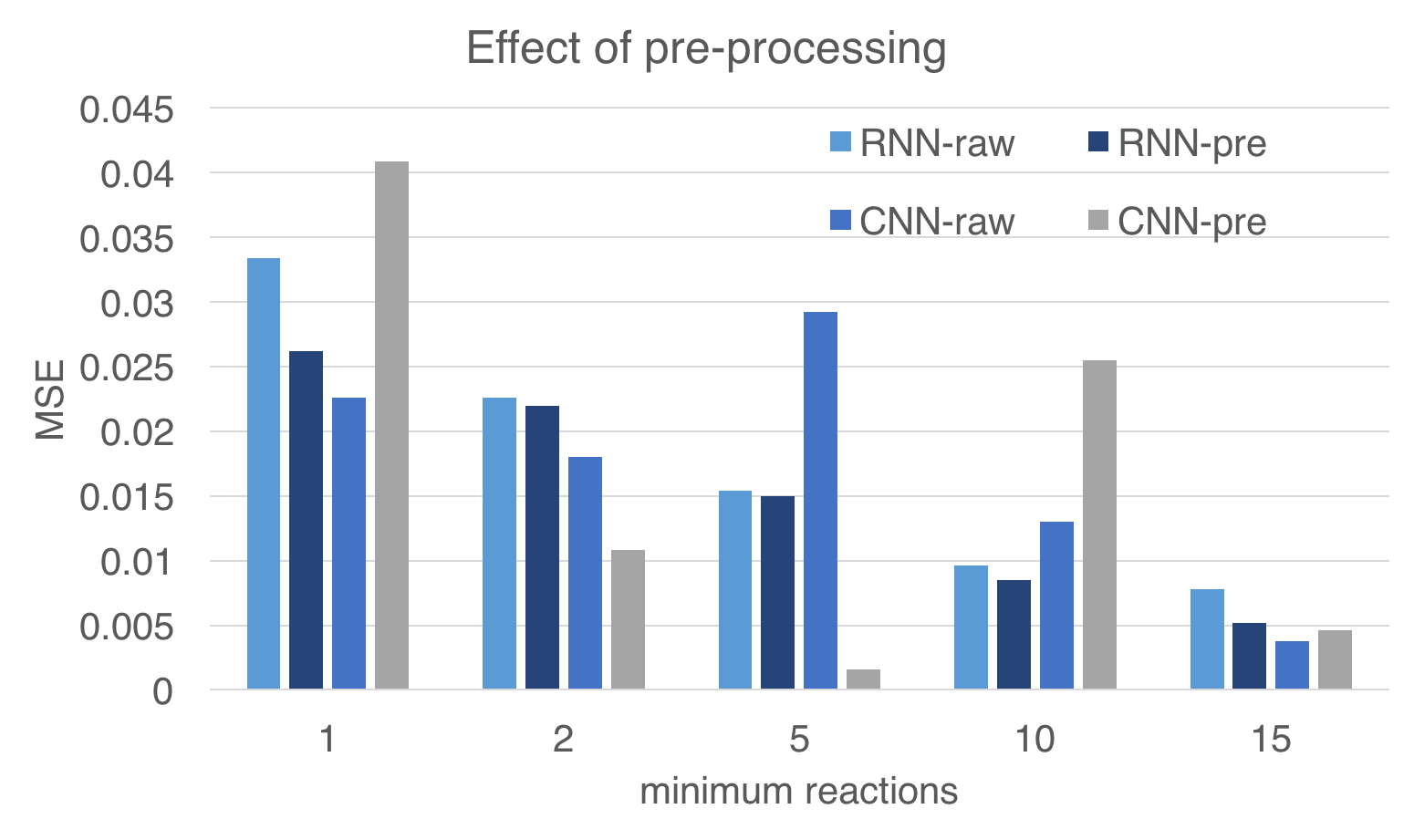}
        \caption{Effect of pre-processing on different models}
        \label{fig:preproc}
\end{figure}

In order to assess the effect of pre-processing on the quality of the trained models, two versions for each neural network were trained. One instance was trained without pre-processing the dataset and the other instance was trained with the pre-processed dataset. Results are cross-validated and here the average values are reported. Figure \ref{fig:preproc} indicates that overall the error was decreasing or being close to equal (which is applicable for both CNN and RNN). The x-axis represents the minimum number of `non-like' reactions in order to be included in the dataset. It should be noted that these models were trained on the basis of having 6 outputs (one for each reaction), thus the result might be affected by the skewed distribution over many `like' reactions. This is the reason that the pre-processed version of CNN performs very well for posts with 5 minimum reactions and very bad for posts with 10 minimum reactions In addition, the variance for the different cross-validation results was high. In the next subsection we explore what happens after the removal of `like' reactions.

\subsection{Exclusion of like reactions}\label{sec:exLikes}
\begin{figure}[h]
          \includegraphics[width=\columnwidth]{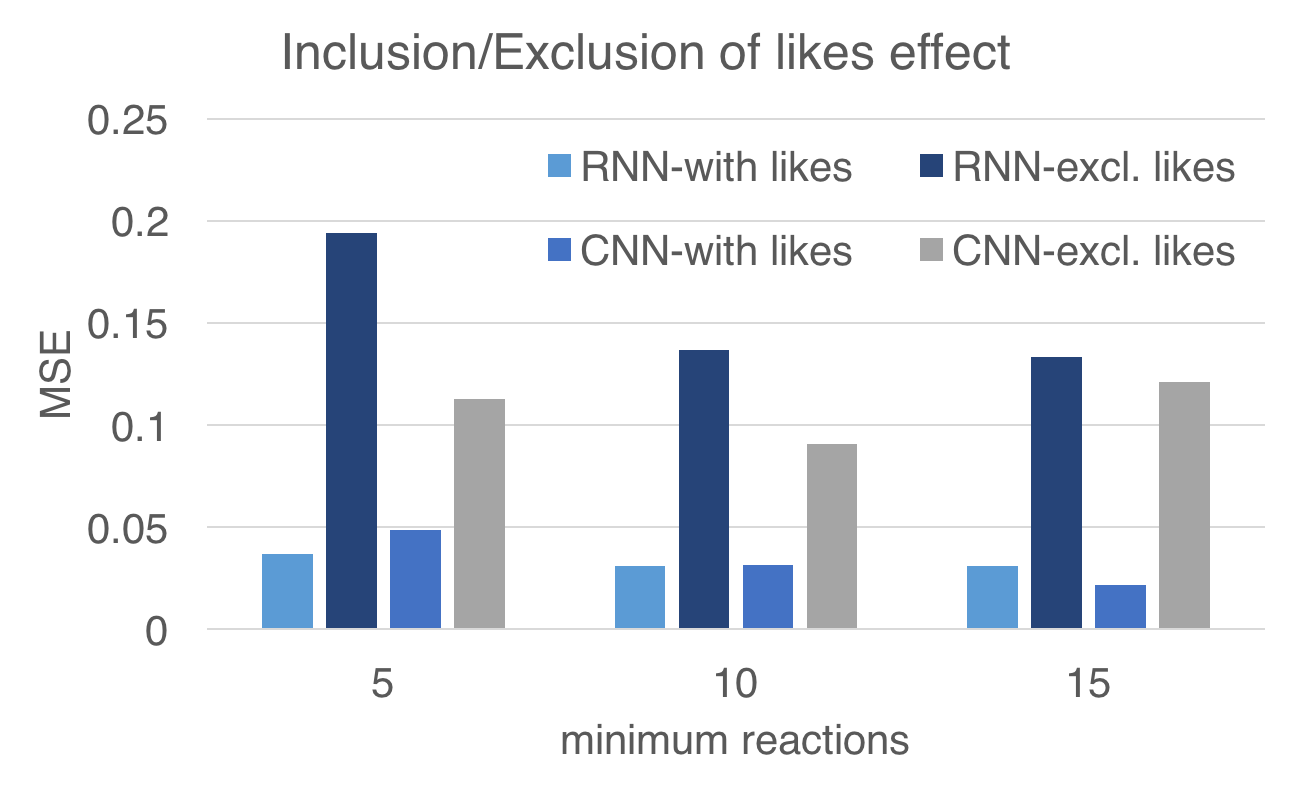}
        \caption{Effect of inclusion/exclusion of likes on different models}
        \label{fig:nolikes}
\end{figure}

Early results showed that including the original \textit{like} reaction in the models would lead to meaningless results. The huge imbalanced dataset led to predicting a 100\% ratio for the \textit{like} reaction. In order to tackle this issue, the \textit{like} reactions are not fed into the models during the training phase (moreover the \textit{love} reaction can be used for equivalent purposes, since they express similar emotions). Figure \ref{fig:nolikes} shows an increase of the error when the likes are ignored. The explanation for this increase is related to heavily unbalanced distribution of \textit{like} reactions: Although there is an increase in the error, predictions now are more meaningful than always predicting a like ratio close to 100\%. After all, it is the relative reaction distribution that we are interested in predicting.

\subsection{Ensemble performance}\label{sec:mse}
Table \ref{tab:networks} summarizes the testing error for the CNN and RNN with respect to the same split dataset and by also taking the validation error into account. One can see that RNN performs better than CNN, although it requires additional training time. Results are cross-validated on 10 different runs and variances are presented in the Table as well.

\begin{table}[h]
	\centering
	\caption{RNN and CNN comparison after cross-validation}
	\scalebox{1}{
		\begin{tabular}{c|c|c}
			\hline
			& \multicolumn{1}{l}{MSE} & \multicolumn{1}{l}{\# Epochs} \\
			\hline
			CNN   & 0.186 ($\pm 0.023$) & 81 \\
			\hline
			RNN   & 0.159 ($\pm 0.017$) & 111 \\
			\hline
	\end{tabular}}
	\label{tab:networks}
\end{table}

\begin{figure}[h]
	    \centering
	    \includegraphics[width = \columnwidth]{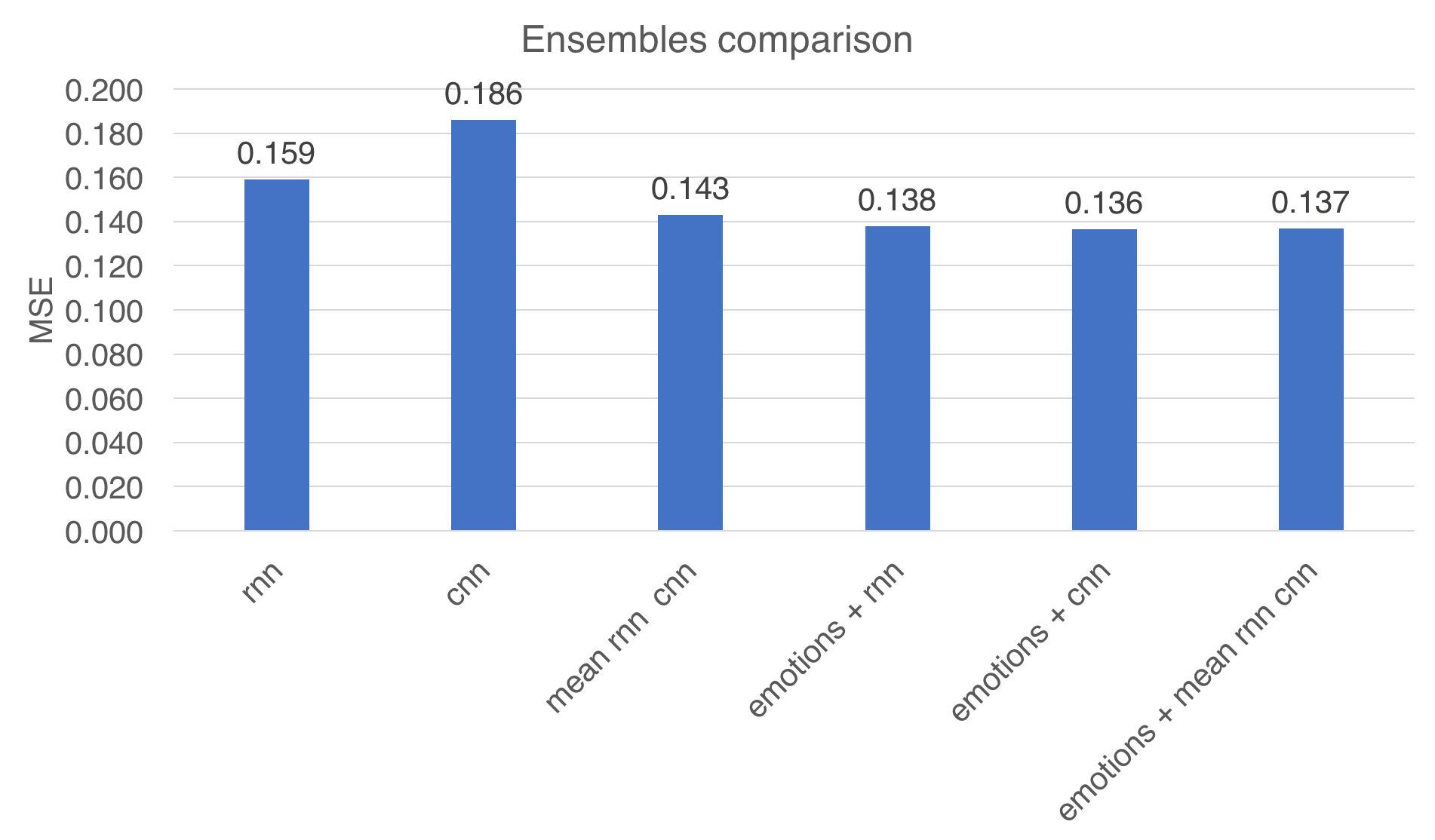}
	    \caption{Performance results for different combinations of the neural networks and emotions.}
	    \label{fig:predictresult}
    \end{figure}

Combined results for either of the networks and the emotion miner can be seen in Figure \ref{fig:predictresult}. The networks themselves have the worst results but an average combination of both is able to achieve a better result. Optimal result is achieved by the emotions + cnn combination, although this difference is not significant than other combinations. These results can be boosted by optimizing the hyperparameters of the networks and also by varying different amount of posts. As a conclusion one can say that using emotions to combine them with neural network output improves the results of prediction. 

\begin{figure*}[!h]
\centering
\includegraphics[width = \textwidth]{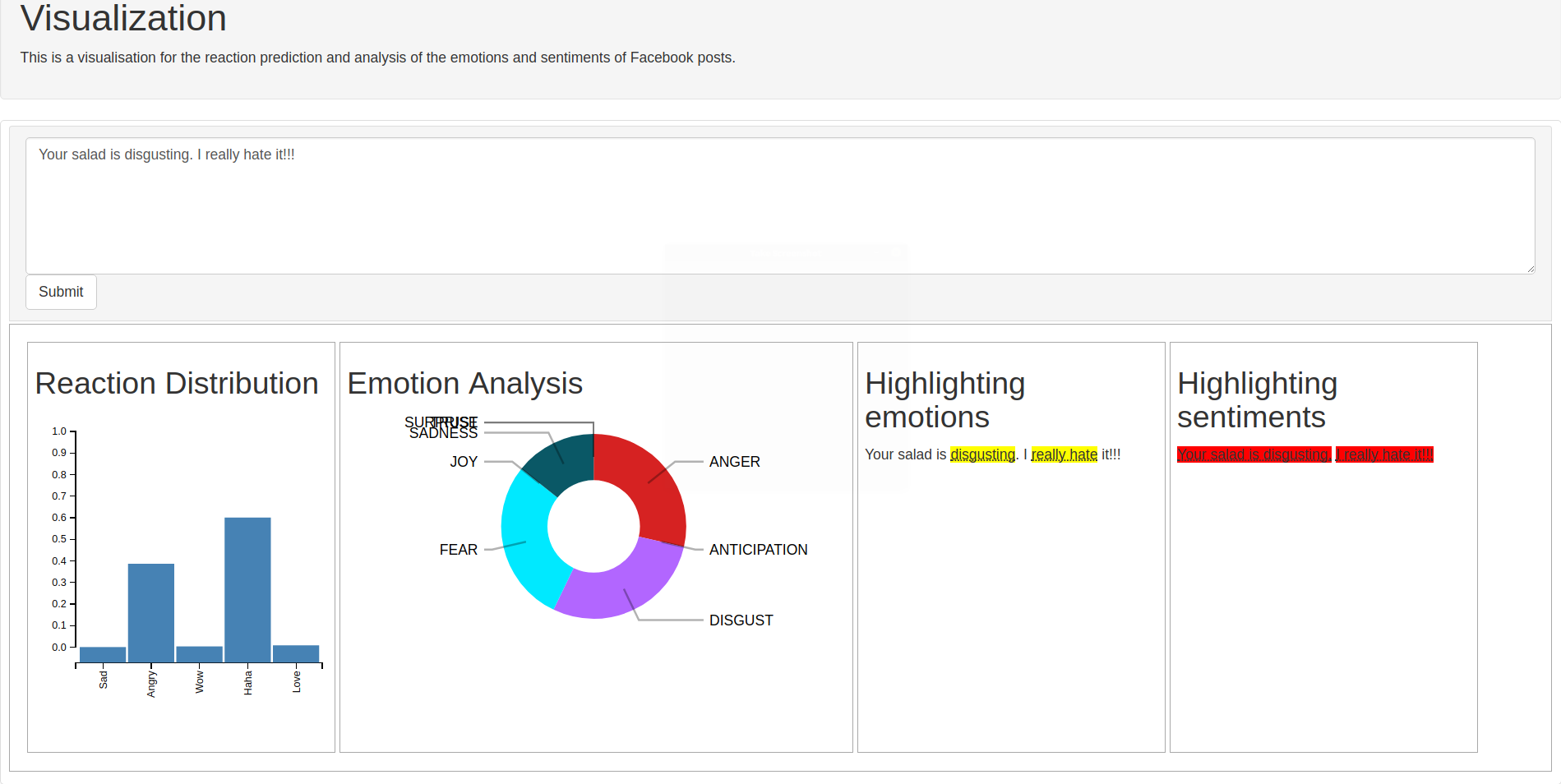}
	\caption{Example visualisation}
	 \label{fig:vis}
\end{figure*}

Finally, we present a simple, yet effective visualization environment which highlights the results of the current paper, that can be found in Figure \ref{fig:vis}. In this figure, one can see at the input field of the Facebook post on the top and then four different result panels: the first one shows the reaction distribution, the second panel shows the proportions of the eight emotions, the third panel highlights the emotions (and by hovering one can see the total shows the overall distribution (vector of eight) and the fourth panel shows the highlighting of the sentiments.
	
\section{Conclusion}
\label{sec:conclusion}
In this paper, a framework for predicting the Facebook post reaction distribution was presented, trained on a customer service dataset from several supermarket Facebook posts. This study revealed that a baseline sentiment miner can be used in order to detect a post sentiment/emotion. Afterwards, these results can be combined with the output of neural network models to predict the Facebook reactions. While there has been a lot of research around sentiment analysis, emotion mining is still mostly uncharted territory and this work also contributes to this direction. The used dataset is available for other researchers and can be also used as a baseline for performing further experiments. In addition, a more accurate evaluation of the emotion miner can be conducted by using the MPQA corpus \cite{mpqa}.

Facebook reaction predictions can clearly enhance customer experience analytics. Most companies are drowned in social media posts, thus a system that identifies the emotion/reaction prediction of a post in almost real-time can be used to provide effective and useful feedback to customers and improve their experience. So far in the dataset, the reaction of the page owner has not been included but this could be useful information on how the post was addressed (or could be addressed).

Future work includes working towards refining the architectures of the neural networks used. Moreover, one of the next steps is to implement a network that predicts the (absolute) amount of reactions (and not just the ratio). This number is of course susceptible to external parameters (e.g. popularity of the post/poster, inclusion of other media like images or external links, etc.), so another direction would be to include this information as well. More specifically, the combination of images and text can reveal possible synergies in the vision and language domains for sentiment/emotion related tasks.

\bibliographystyle{apalike}
{\small
\bibliography{references.bib}}

\end{document}